\documentclass[runningheads,a4paper]{llncs}

\usepackage[cmex10]{amsmath}
\usepackage{amssymb}
\usepackage{llncsdoc}
\usepackage{mathptmx}       
\usepackage{helvet}         
\usepackage{courier}  
\usepackage{float}
\usepackage{xcolor}
\usepackage{soul}
\definecolor{pastelred}{RGB}{255,204,204}
\definecolor{pastelblue}{RGB}{204,229,255}
\definecolor{pastelgreen}{RGB}{204,255,204}
\usepackage{amsmath}
\usepackage{tcolorbox}
\usepackage{makeidx}         
\usepackage[bottom]{footmisc}
\usepackage{flafter} 
\usepackage{url}
\usepackage[linesnumbered,ruled,vlined]{algorithm2e}
\usepackage{array,cite}
\usepackage{booktabs}
\usepackage[caption=false,font=footnotesize]{subfig}
\usepackage{xcolor,colortbl}
\usepackage{bm}
\usepackage{subfig}
\usepackage{placeins}
\usepackage{multirow}

\DeclareMathSymbol{\R}{\mathalpha}{AMSb}{"52}

\definecolor{Gray}{gray}{0.85}
\definecolor{LightCyan}{rgb}{0.88,1,1}

\begin{document}
\mainmatter 

\title{TACE: Tumor-Aware Counterfactual Explanations}

\author{Eleonora Beatrice Rossi \and Eleonora Lopez \and Danilo Comminiello
\thanks{
This work was supported by the European Union under the National Plan for Complementary Investments to the Italian National Recovery and Resilience Plan (NRRP) of NextGenerationEU,  Project PNC 0000001 D3 4 Health, (Digital Driven Diagnostics, prognostics and therapeutics for sustainable Health care) - SPOKE 1 Clinical use cases and new models of care supported by AI/E-Health based solutions - CUP B53C22006120001. This work was also supported by the Italian Ministry of University and Research (MUR) within the PRIN 2022 Program for the project ``EXEGETE: Explainable Generative Deep Learning Methods for Medical Signal and Image Processing", under grant number 2022ENK9LS, CUP C53D23003650001.
}}
\titlerunning{TACE: Tumor-Aware Counterfactual Explanations}
\authorrunning{Rossi et al.}
\institute{Department of Information Engineering, Electronics and Telecommunications (DIET), \\ ``Sapienza'' University of Rome, \\ Via Eudossiana 18, 00184, Rome. \\Email: elebearossi@gmail.com, eleonora.lopez@uniroma1.it}

\maketitle

\begin{abstract}
The application of deep learning in medical imaging has significantly advanced diagnostic capabilities, enhancing both accuracy and efficiency. Despite these benefits, the lack of transparency in these AI models, often termed ``black boxes," raises concerns about their reliability in clinical settings. Explainable AI (XAI) aims to mitigate these concerns by developing methods that make AI decisions understandable and trustworthy. In this study, we propose Tumor Aware Counterfactual Explanations (TACE), a framework designed to generate reliable counterfactual explanations for medical images. Unlike existing methods, TACE focuses on modifying tumor-specific features without altering the overall organ structure, ensuring the faithfulness of the counterfactuals. We achieve this by including an additional step in the generation process which allows to modify only the region of interest (ROI), thus yielding more reliable counterfactuals as the rest of the organ remains unchanged. We evaluate our method on mammography images and brain MRI. We find that our method far exceeds existing state-of-the-art techniques in quality, faithfulness, and generation speed of counterfactuals. Indeed, more faithful explanations lead to a significant improvement in classification success rates, with a 10.69\% increase for breast cancer and a 98.02\% increase for brain tumors. The code of our work is available at \url{https://github.com/ispamm/TACE}.

\keywords{Generative counterfactual explanations, explainability, medical images, breast cancer, brain MRI}
\end{abstract}

\section{Introduction}
\label{sec:introduction}
The progress of deep learning and artificial intelligence in medical image analysis has significantly enhanced the diagnosis process, improving both accuracy and efficiency. However, the opaque nature of these architectures often referred to as "black boxes", presents crucial challenges for their reliability and transparency, which are fundamental aspects of medical applications \cite{rasheed2022explainable}. Furthermore, as advancements in the field continue, the models become increasingly complex and, consequently, more difficult to interpret. Explainable AI (XAI) is a field of research that addresses this need and, for this reason, has gained increasing attention in recent years. XAI promotes the development of methods that allow users to understand and trust the results produced by AI, ensuring that they can transparently communicate the origin of their decisions and simultaneously facilitate the identification of errors and biases.

Counterfactual explainability represents a branch of the research field of XAI. Operating on a causal principle, these explanations propose scenarios such as ``If X had not occurred, Y would not have occurred". In other words, a counterfactual explanation is defined as the smallest semantically meaningful change of a factual input such that a target model changes its prediction.
In medical imaging, where accurate diagnosis and patient safety are fundamental, counterfactual explanations prove to be an effective choice because they allow healthcare professionals to see how slight variations in medical images could lead to different diagnostic outcomes by providing clear and understandable examples. Nevertheless, existing methods of counterfactual explainability for medical imaging present a recurrent issue of not only modifying the region of interest (ROI) but also changing the shape of the organ itself. This is a problem because, first, it does not respect the definition of a counterfactual, and, most importantly, in this way, the output counterfactual is not reliable.

In this paper, we address these challenges and develop a framework for reliable counterfactual generation of medical images, i.e., Tumor Aware Counterfactual Explanations (TACE). We achieve this with two main features. First, we make the generation process ``tumor-aware" by focusing on identifying and manipulating tumor characteristics such as size, shape, and intensity. Moreover, we include an additional step in the generation process which allows to modify only the ROI, thus yielding more reliable counterfactuals as the rest of the organ remains unchanged. In this manner, our method is able to generate highly accurate and faithful counterfactual explanations that are particularly beneficial for clinicians. Moreover, these tumor-aware explanations provide critical insights into the features that lead a diagnostic AI model to classify a region as tumorous, aiding in the deciphering and trusting of the AI’s decision-making process, as well as in their training and validation.

We evaluate our method on two different medical problems. For breast cancer classification from mammography images we utilize the union of two publicly available datasets, mini-DDSM and INbreast. For brain tumor classification from MRI data, we combine three public datasets available on Kaggle. We compare our approach to a state-of-the-art method, StylEx, and find that TACE far exceeds it in terms of both quality of counterfactuals as well as generation speed. Indeed, the counterfactual explanations generated with our method result much more faithful to the original factual image compared to StylEx, yielding a classification success-rate higher by a factor of 10.69\% for breast cancer and 98.02\% for brain MRI.

\section{Related works}
\label{sec:related}
In medical imaging, counterfactual explanations have shown significant promise. Neal et al. \cite{neal2018open} utilized these methods for data augmentation in open-set learning tasks. Wang et al. \cite{wang2021bilateral} applied counterfactuals to improve mammogram classification, leveraging anatomical symmetry in human breasts. However, their generation algorithm inherently relies on the symmetry of body parts, strongly limiting the generalization capabilities of their approach. Zhao et al. \cite{zhao2020fast} proposed using a StarGAN architecture to create counterfactual explanation images. However, the system was only applied to binary images, i.e., images where each pixel is either black or white. The resulting counterfactuals were used to highlight the pixels that differ between original and counterfactual images. One of the most recent and significant studies that we consider as a benchmark is StylEx \cite{lang2021explaining}, which drawing from previous works \cite{singla2019explanation, goetschalckx2019ganalyze, singla2023explaining, antoran2020getting}, has become a state-of-the-art approach to the explainability of deep learning models, particularly in generating counterfactual explanations. It has been tested on various types of data, including medical images, demonstrating its versatility. In the original paper, it was applied to retinal images, and subsequent studies like \cite{atad2022chexplaining} have further explored its potential in the medical domain. However, StylEx may not always provide clear, actionable insights into the causative factors behind these changes. One significant issue observed in the medical domain is that StylEx often modifies the shape of the organ itself when generating counterfactual explanations. This alteration does not respect the definition of a counterfactual, which should involve the smallest semantically meaningful change. As a result, the output counterfactuals generated by StylEx may not be reliable or interpretable, as they can introduce unrealistic anatomical variations that do not correspond to real medical conditions. These limitations highlight the need for more specialized approaches, such as our proposed Tumor Aware Counterfactual Explanations (TACE), which focuses on maintaining the anatomical integrity of the region of interest (ROI) while providing semantically meaningful explanations. 

\section{Proposed method}
\label{sec:method}

\begin{figure*}[t]
    \centering
    \includegraphics[width=\textwidth]{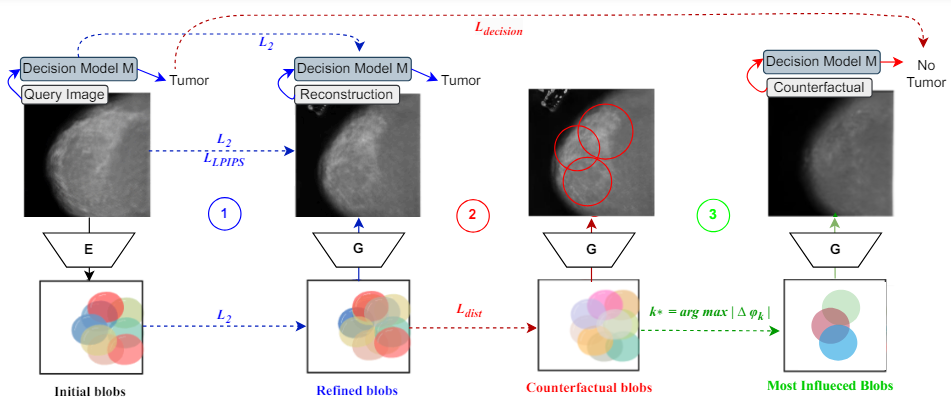}
    \caption{The TACE framework involves three main steps. First, the query image is processed by the encoder (E), which converts it into a set of initial blobs representing various spatial and style parameters. Next, the generative network (G) refines these initial blobs to produce a reconstruction of the query image. Finally, the refined blobs are modified to generate counterfactual explanations. The blobs with the most significant changes are identified and adjusted to maintain high similarity to the original image, providing clear and meaningful counterfactual explanations.}
    \label{fig:method}
\end{figure*}

We propose Tumor-Aware Counterfactual Explanations (TACE), a novel framework designed to provide counterfactual explanations specifically for medical imaging, depicted in Fig.~\ref{fig:method}. TACE is inspired by OCTET \cite{zemni2023octet}, a framework originally designed for generating counterfactual explanations in autonomous driving scenarios. OCTET leverages an encoder and inversion process to modify specific elements within an image to create counterfactual examples. Unlike StylEx \cite{lang2021explaining}, this method employs as backbone generative model a BlobGAN \cite{epstein2022blobgan}. This network is more suited to counterfactual generation as it leverages a latent space defined by a set of ``blobs", i.e., latent codes $\phi$ and $\psi$ representing spatial and style parameters respectively, in order to generate images. In this work, we have adapted and modified OCTET \cite{zemni2023octet} to suit the unique requirements of medical imaging and we have introduced an additional generation step to enhance the fidelity of the generated counterfactuals. The TACE process begins with an encoder network $E$ that transforms input images into a structured latent space defined by a set of parameters representing various objects in the scene, i.e., blobs. The encoder is trained in two phases, a pre-training and a fine-tuning phase. In the pre-training phase, the encoder $E$ is trained using generated images from a generative network $G$ to align its output $\mathbf{h}$ with the generative network through the following reconstruction objective:
\begin{equation}
L^\text{pretrain}_{E} = L_{2}(\mathbf{h}, E(G(\mathbf{h})))
\end{equation}
In this formula the term \(E(G(\mathbf{h}))\) represents the process where the image generated by \(G\) from the latent vector \(\mathbf{h}\) is encoded back into the latent space by \(E\). The \(L_{2}\) term denotes the L2 loss, which measures the difference between the original latent vector \(\mathbf{h}\) and the reconstructed latent vector \(E(G(\mathbf{h}))\). In the second phase, the encoder is fine-tuned on both real and generated images with a composite objective that includes terms for reconstruction accuracy, perceptual similarity (LPIPS loss), and alignment with the decision model $f_M$:
\vspace{-0,2cm}
\begin{equation}
\begin{split}
L^\text{finetune}_{E} =  & \colorbox{pastelred}{$\displaystyle L_{2}(\mathbf{x}, G(E(\mathbf{x}))) + \lambda_\text{LPIPS} L_\text{LPIPS}(\mathbf{x}, G(E(\mathbf{x})))$} +\\
& \colorbox{pastelblue}{$\displaystyle \lambda_\text{latent} L_{2}(\mathbf{h}, E(G(\mathbf{h})))$} + \colorbox{pastelgreen}{$\displaystyle \lambda_\text{decision} L_{2}(f_M(\mathbf{x}), f_M(G(E(\mathbf{x}))))$}
\end{split}
\end{equation}

\vspace{-0,06cm}
\noindent This formula emphasizes the importance of three key aspects:
\vspace{-0,2cm}
\begin{itemize}
  \item \textbf{Image Reconstruction Accuracy}: Images encoded as latent parameters must be accurately reconstructed by the generator. This is achieved by employing a combination of the perceptual  \sethlcolor{pastelred}\hl{$L_\text{LPIPS}$} loss and the \sethlcolor{pastelred}\hl{$L_\text{2}$} loss, ensuring the faithfulness of the reconstruction.
  
  \item \textbf{Re-encoding Generated Images}: The method includes a process for re-encoding generated images back into the latent space. This step is crucial for keeping the latent parameters within the generator domain, which is achieved using the \sethlcolor{pastelblue}\hl{$L_{2}$} loss. loss.
  
  \item \textbf{Consistency with Decision Model}: A key component of this approach is the focus on ensuring that the reconstructed images are consistent with the decision model. This is measured through the \sethlcolor{pastelgreen}\hl{$L_\text{2}$} distance between the features extracted from both the original and reconstructed images. The aim here is to ensure that the reconstructed images are not only realistic but also faithful to the original decisions made by the model.
\end{itemize}
\vspace{-0,07cm}
Another critical step of TACE is the inversion process, which involves obtaining a latent code corresponding to the query image using a BlobGAN generator backbone. Given a query image denoted by \(\mathbf{x}^q\), the objective is to find the corresponding latent codes \((\phi^q, \psi^q)\), i.e., the blobs, such that the generator \(G\) can recover the query image \(\mathbf{x}^q\) as accurately as possible. This process ensures that the generated image closely matches the query image and retains key features relevant to the model decision, even if there is more than one tumor. The optimization objective is the following:

\begin{equation}
\begin{split}
\phi^q, \psi^q = \arg \min_{\phi, \psi} \hspace{0.2cm}&L_{\text{LPIPS}}(G(\phi, \psi), \mathbf{x}^q) + L_{2}(G(\phi, \psi), \mathbf{x}^q) \\
&+ L_2(f_M(\mathbf{x}^q), f_M(G(\phi, \psi))) + L_{2}(\phi, \psi, E(\mathbf{x}^q)).
\end{split}
\label{eq:linversion}
\end{equation}
This optimization equation is composed of four terms. The first term, $L_\text{LPIPS}$, employs a perceptual loss to ensure that the generated image from the blob parameters closely reconstructs the query image $\mathbf{x}^q$. The second term, an $L_\text{2}$ loss, enhances the accuracy of this reconstruction. Then, the third term introduces an $L_\text{2}$ loss on intermediate features of the decision model $f_M$, aimed at preserving the features significant to the model decision-making process. Finally, the fourth term encourages the parameters to remain close to the initial solution provided by the encoder, ensuring a more accurate inversion.

While OCTET is effective for scenarios involving multiple objects, such as autonomous driving, it is not directly suitable for medical imaging. Indeed, by employing the OCTET method directly on medical images we have found several issues. As it was designed for autonomous driving, when applied to medical images the method tends to generate counterfactuals that change the overall shape of the organ and, thus, are not faithful. To address this limitation, TACE introduces an additional intermediate process during the optimization of the blobs. It is designed to increase the fidelity between the counterfactual images and the original images by focusing on the most significant changes. Specifically, the blob that undergoes the most significant changes during optimization is identified. Then, an image is generated focusing solely on this blob and its neighboring blobs, which are modified to increase their resemblance to the original image. This process ensures that the counterfactual image maintains a high degree of similarity to the original, thereby avoiding errors and providing more accurate and meaningful explanations. In detail, the steps are the following:
\begin{enumerate}
    \item Identify the blob \( \phi_k \) with the most significant change during optimization:
    \[
    k^* = \arg \max_{k} \|\Delta \phi_k\|
    \]
    where \( \Delta \phi_k \) refers to the variation in the parameters of blob \( \phi_k \).

    \item Focus on the blob \( \phi_{k^*} \) and its neighboring blobs \( \mathcal{N}(k^*) \). The neighboring blobs are defined as those that intersect the area of the blob initially found, i.e., all blobs that overlap with the region associated with \( \phi_{k^*} \).
\end{enumerate}
Thus, our final objective becomes:
\begin{equation}
\begin{split}
\phi^q, \psi^q = \arg \min_{\phi, \psi} \hspace{0.2cm} &L_{\text{LPIPS}}(G(\phi, \psi), \mathbf{x}^q) + L_{2}(G(\phi, \psi), \mathbf{x}^q) \\
&+ L_2(f_M(\textbf{x}^q), f_M(G(\phi, \psi))) + L_{2}(\phi, \psi, E(\textbf{x}^q)) \\
&\text{subject to:} \\
& \phi = \{\phi_{k^*}, \phi_{\mathcal{N}(k^*)}\},
\end{split}
\label{eq:finalobjective_simplified}
\end{equation}
The additional step in TACE identifies the blob that most significantly influences the classifier during the optimization process. By focusing on modifying this blob and its peripheral blobs, TACE generates counterfactual explanations that are more faithful to the original images. This approach ensures that the generated counterfactuals do not introduce irrelevant or misleading features. 
Unlike previous methods such as StylEx and OCTET (when tested on medical images), TACE modifies only the region of interest (ROI) and does not alter the shape of the organ.


\section{Experimental evaluation}
\label{sec:experiments}

\begin{table}[t]
\centering
\caption{Generation quality of backbone generative models. Values in bold indicate the best method.}
\label{tab:backbone}
\begin{tabular}{@{}llllclcc@{}}
\toprule
\textbf{Dataset}           &  & \textbf{Backbone} &  & \textbf{Steps}           &  & \textbf{FID $\downarrow$} & \textbf{LPIPS $\downarrow$} \\ \midrule
\multirow{2}{*}{Brain MRI} &  & StyleGAN          &  & 108K                     &  & 81.96                     & 0.46                        \\
                           &  & BlobGAN           &  & 108K                     &  & \textbf{81.70}            & \textbf{0.37}               \\ \midrule
\multirow{2}{*}{Mammogram} &  & StyleGAN          &  & \multicolumn{1}{l}{105K} &  & 55.80                     & 0.57                        \\
                           &  & BlobGAN           &  & \multicolumn{1}{l}{105K} &  & \textbf{55.02}            & \textbf{0.44}               \\ \bottomrule
\end{tabular}
\end{table}

\begin{table}[t]
\centering
\caption{Evaluation of counterfactual quality for both mammogram and brain MRI datasets in terms of Success Rate (SR), LPIPS and FID. Values in bold and underlined indicate best and second best, respectively.}
\begin{tabular}{llllccc}
\toprule
\textbf{Dataset} & & \textbf{Method} & & \textbf{SR (\%) $\uparrow$}  & \textbf{FID $\downarrow$} & \textbf{LPIPS $\downarrow$} \\
\midrule
\multirow{3}{*}{Brain MRI} & & StylEx & & 42.67 & 0.76 & 0.33  \\
& & OCTET & & \textbf{94.66} & \underline{0.72} & \underline{0.26}  \\
& & TACE & & \underline{84.49} & \textbf{0.54} & \textbf{0.13}  \\
\midrule
\multirow{3}{*}{Mammogram} & & StylEx & & 74.67 & 0.88 & 0.74  \\
& & OCTET & & \textbf{99.33} & \underline{0.80} & \underline{0.33}  \\
& & TACE & & \underline{82.65} & \textbf{0.78} & \textbf{0.20}  \\
\bottomrule
\end{tabular}
\label{tab:comparison}
\end{table}

\subsection{Datasets}
For this study, we focus on brain tumors using MRI images and breast cancer using mammography images. 
\noindent The Brain Tumor MRI dataset includes 7023 images from Figshare, SARTAJ, and Br35H, categorized into glioma, meningioma, pituitary tumor, and no tumor. We use only Normal and Meningioma categories with axial orientation. Data augmentation corrected the initial imbalance, resulting in 300 images for testing, 2683 for training, and 300 for validation in both categories.
\noindent The Mammography dataset combines Mini-DDSM and INbreast images. Only Normal (BI-RADS 1) and Cancer (BI-RADS 4 and 5) categories were used, with craniocaudal (CC) view images. The final dataset includes 120 images for testing, 1146 for training, and 120 for validation in the No Cancer category, and 120 images for testing, 997 for training, and 120 for validation in the Cancer category.

\subsection{Experimental setup}

All experiments are performed using a single NVIDIA V100 GPU, ensuring efficient processing and analysis of the datasets. As classifier network that we aim to explain, we employ a DenseNet121 model trained separately on the brain and mammography datasets with a binary cross-entropy loss, a ReduceLROnPlateau scheduler for learning rate adjustment, and early stopping to prevent overfitting. The classifiers achieve an accuracy of 87\% on the mammography dataset and 88\% on the brain MRI dataset.

We compare our method against two state-of-the-art frameworks, i.e., StylEx \cite{lang2021explaining} and OCTET \cite{zemni2023octet}. For StylEx, the StyleGAN generator is trained conditioned on the DenseNet121 classifier, while for OCTET and TACE, the BlobGAN generator is trained independently of the classifier and configured to generate a fixed number of 20 blobs.  
All generative backbones are trained with the Adam optimizer, a learning rate of 0.001 and a batch size of 8, with the learning rate decayed by a factor of 0.5 every 20.000 steps. For the mammography dataset, the training stops at around 108.000 steps, while for the brain MRI at 105.000 steps. Then, for both OCTET and TACE, following the BlobGAN training, the encoder network is pre-trained on synthetic images generated by BlobGAN and then fine-tuned on real images. The encoder training uses the Adam optimizer with a learning rate of 0.002, and the fine-tuning phase continues for 15.500 steps.

\subsection{Metrics}


\begin{table}[t]
\centering
\caption{Efficiency evaluation of the different frameworks. Values in bold indicate the fastest execution time.}
\begin{tabular}{llccc}
\toprule
\textbf{Method} & & \textbf{\#Counterfactuals} & & \textbf{Time (hours)} \\
\midrule
StylEx & & 250 & & 30,5 \\
OCTET & & 250 & & \textbf{2.5} \\
TACE & & 250 & & \textbf{2.5} \\
\bottomrule
\end{tabular}
\label{tab:execution_time_comparison}
\end{table}

\begin{figure}[t]
    \centering
        \includegraphics[width=0.8\textwidth]{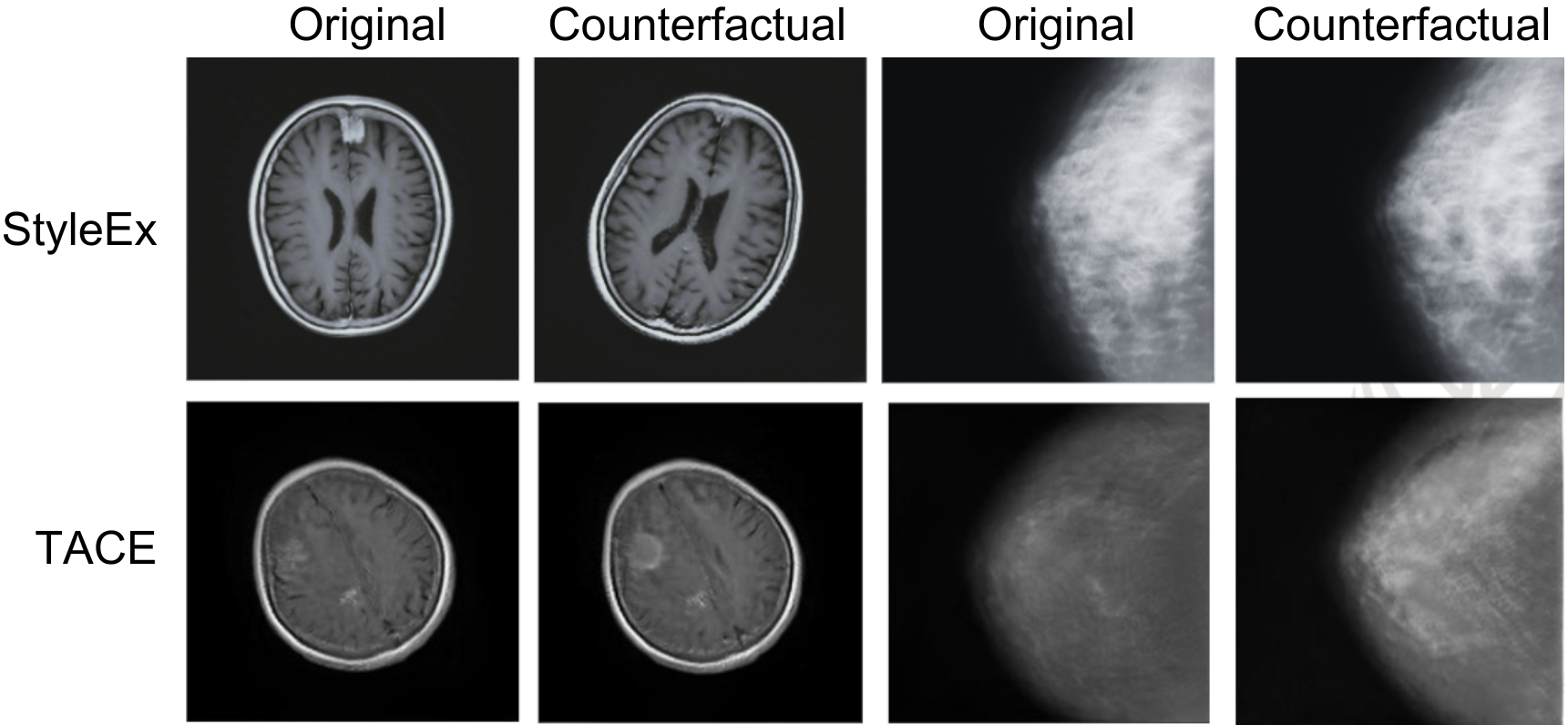}
        \caption{Counterfactuals of brain MRI generated with StylEx and with TACE. It is clear that StylEx modifies the colors, the internal structure of the brain/breast, and its orientation. In contrast, TACE generates faithful counterfactuals.}
        \label{fig:stylex_failure1}
\end{figure}




\begin{figure}[t]
    \centering
    \includegraphics[width=0.6\linewidth]{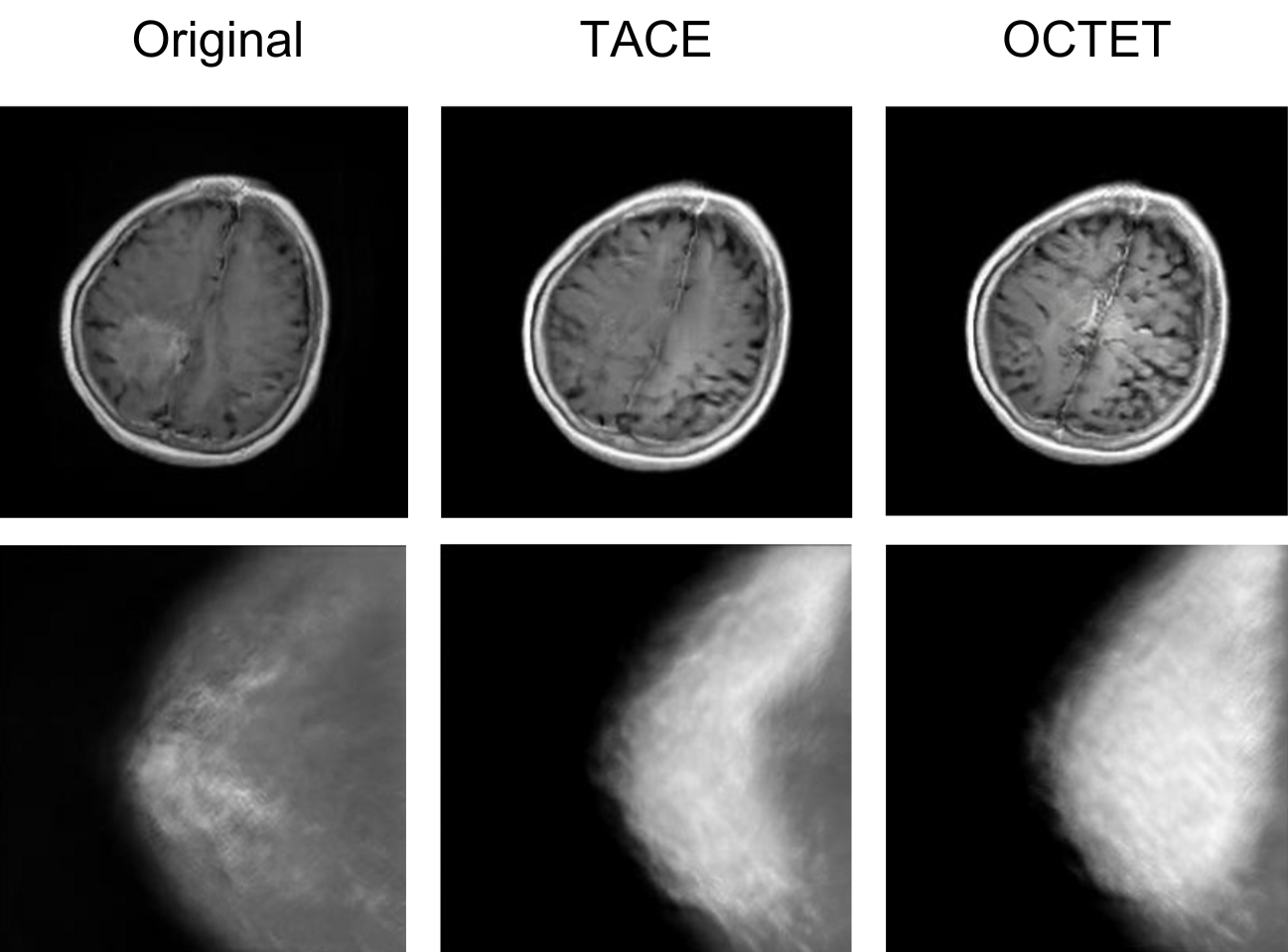}
    \caption{Comparison of OCTET and TACE. In brain MRIs, OCTET modifies the internal brain structure, adding details not present in the original image. In contrast, TACE limits modifications to the tumor area, making the changes more imperceptible and targeted. In mammographies, the tumor area added by OCTET covers nearly the entire breast, whereas TACE modifications are much more circumscribed and precise.}
    \label{fig:octet_vs_tace2}
\end{figure}

For the evaluation of the generative networks, i.e., BlobGAN and StyleGAN, across all three methods, i.e., StylEx, OCTET, and TACE, we use Fréchet Inception Distance (FID) and Learned Perceptual Image Patch Similarity (LPIPS). These metrics assess the quality and fidelity of generated images by comparing them to real images. Lower FID and LPIPS values indicate higher quality and closer perceptual similarity to real images. For evaluating counterfactual explanations, we also use LPIPS and FID to ensure quality and fidelity. Additionally, we measure the Success Rate, which is the percentage of counterfactuals that successfully change the classifier decision, following the approach of \cite{zemni2023octet}. Higher success rates indicate more effective counterfactual generation. Finally, we measure the time required to generate 250 counterfactuals to assess the computational efficiency of each method.

\subsection{Experimental results}

We perform an extensive experimental evaluation conducting both a quantitative and qualitative analysis on two different medical problems and medical imaging techniques, i.e., brain tumor from MRI images and breast cancer from mammography images.

\subsubsection{Quantitative analysis}

Firstly, we analyze the generation quality of the different backbone models. As shown in Tab.~\ref{tab:backbone}, BlobGAN achieves lower FID and LPIPS scores compared to StyleGAN for both datasets. These results indicate that BlobGAN generates images that are closer to real images in terms of both visual quality and perceptual similarity, making it a superior generative network compared to StyleGAN for the evaluated tasks. 

Secondly, we evaluate the quality of the generated counterfactuals and report the results in Tab.~\ref{tab:comparison}. Our method, TACE, focusing only on the region of interest, achieves lower LPIPS and FID scores. Indeed, comparing OCTET and TACE we can observe a slight trade-off between the success rate and the reliability of the generated counterfactuals. However, TACE is still significantly higher than that of StylEx in terms of success rate. A significant increase in the success rate is observed particularly for brain images, which are more complex and diverse. 

Finally, we evaluate the frameworks in terms of efficiency in Tab.~\ref{tab:execution_time_comparison}. Our method, TACE, demonstrates a great superior efficiency in generating counterfactuals. TACE and OCTET both generate 250 counterfactuals in only 2.5 hours, whereas StylEx takes significantly longer, requiring 30.5 hours for the same number of counterfactuals. This demonstrates that TACE is not only more effective in terms of quality metrics but also vastly more efficient in terms of execution time, making it a practical choice for real-world medical imaging applications.



\subsubsection{Qualitative analysis}

We also perform a qualitative analysis to further understand the strengths and limitations of the methods evaluated. StylEx often modifies the shape and orientation of organs and lacks precision in the details, as shown in the examples of Fig.~\ref{fig:stylex_failure1}. Although OCTET performs better than StylEx in quantitative terms, it sometimes also tends to overly modify the original image, as can be seen in Fig.~\ref{fig:octet_vs_tace2}, where we directly compare OCTET and TACE. It is evident that TACE generates counterfactuals that are more faithful to the original images. Indeed, TACE focuses on modifying only the ROI, preserving the overall anatomical structure and ensuring that the counterfactuals are realistic and clinically relevant.

\section{Conclusions}
\label{sec:conclusions}
We propose Tumor Aware Counterfactual Explanations (TACE), a method that generates reliable counterfactual explanations for medical images by focusing on tumor-specific features without altering the overall organ structure. TACE addresses the limitations of existing methods, such as the reliance on anatomical symmetry or the modification of the entire organ shape, by ensuring that only the region of interest (ROI) is modified. The intermediate step introduced in TACE ensures that the generated counterfactuals maintain a high degree of similarity to the original images, providing more accurate and meaningful explanations. Our method was evaluated on breast cancer classification from mammography images and brain tumor classification from MRI data. Both quantitative and qualitative experiments demonstrate that TACE outperforms the state-of-the-art method, StylEx, in terms of counterfactual quality and generation speed. Specifically, TACE improves classification success rates by 10.69\% for breast cancer and 98.02\% for brain tumors. Future work will explore the use of diffusion models to further enhance counterfactual generation.
\bibliographystyle{splncs03}
\bibliography{refs}

\begin{thebibliography}{10}
\providecommand{\url}[1]{\texttt{#1}}
\providecommand{\urlprefix}{URL }

\bibitem{antoran2020getting}
Antorán, J., Bhatt, U., Adel, T., Weller, A., Hernández-Lobato, J.M.: Getting a clue: A method for explaining uncertainty estimates. arXiv preprint arXiv:2006.06848  (2020)

\bibitem{atad2022chexplaining}
Atad, M., Dmytrenko, V., Li, Y., Zhang, X., Keicher, M., Kirschke, J., Wiestler, B., Khakzar, A., Navab, N.: Che{X}plaining in style: Counterfactual explanations for chest {X}-rays using {S}tyle{GAN}. arXiv arXiv:2207.07553  (2022)

\bibitem{epstein2022blobgan}
Epstein, D., Park, T., Zhang, R., Shechtman, E., Efros, A.A.: {BlobGAN}: Spatially disentangled scene representations. In: European Conference on Computer Vision ({ECCV}). pp. 616--635. Springer (2022)

\bibitem{goetschalckx2019ganalyze}
Goetschalckx, L., Andonian, A., Oliva, A., Isola, P.: {GAN}alyze: Toward visual definitions of cognitive image properties. In: {IEEE}/{CVF} International Conference on Computer Vision ({ICCV}). pp. 5744--5753 (2019)

\bibitem{lang2021explaining}
Lang, O., Gandelsman, Y., Yarom, M., Wald, Y., Elidan, G., Hassidim, A., Freeman, W.T., Isola, P., Globerson, A., Irani, M., et~al.: Explaining in style: Training a {GAN} to explain a classifier in {S}tyle{S}pace. In: {IEEE}/{CVF} International Conference on Computer Vision ({ICCV}). pp. 693--702 (2021)

\bibitem{neal2018open}
Neal, L., Olson, M., Fern, X., Wong, W.K., Li, F.: Open set learning with counterfactual images. In: European Conference on Computer Vision (2018)

\bibitem{rasheed2022explainable}
Rasheed, K., Qayyum, A., Ghaly, M., Al-Fuqaha, A., Razi, A., Qadir, J.: Explainable, trustworthy, and ethical machine learning for healthcare: A survey. Computers in Biology and Medicine  149,  106043 (2022)

\bibitem{singla2023explaining}
Singla, S., Eslami, M., Pollack, B., Wallace, S., Batmanghelich, K.: Explaining the black-box smoothly—a counterfactual approach. Medical Image Analysis  84,  102721 (2023)

\bibitem{singla2019explanation}
Singla, S., Pollack, B., Chen, J., Batmanghelich, K.: Explanation by progressive exaggeration. arXiv preprint arXiv:1911.00483  (2019)

\bibitem{wang2021bilateral}
Wang, C., Li, J., Zhang, F., Sun, X., Dong, H., Yu, Y., Wang, Y.: Bilateral asymmetry guided counterfactual generating network for mammogram classification. IEEE Transactions on Image Processing  30,  7980--7994 (2021)

\bibitem{zemni2023octet}
Zemni, M., Chen, M., Zablocki, {\'E}., Ben-Younes, H., P{\'e}rez, P., Cord, M.: {OCTET}: Object-aware counterfactual explanations. In: {IEEE}/{CVF} Conference on Computer Vision and Pattern Recognition ({CVPR}). pp. 15062--15071 (2023)

\bibitem{zhao2020fast}
Zhao, Y.: Fast real-time counterfactual explanations. arXiv preprint arXiv:2007.05684  (2020)

\end{thebibliography}

\end{document}